\documentclass[journal]{IEEEtran}
\IEEEoverridecommandlockouts
\usepackage[bookmarks=false]{hyperref}
\usepackage{cite}
\usepackage{amsmath,amssymb,amsfonts}
\usepackage{graphicx}
\usepackage{textcomp}
\usepackage{xcolor}

\usepackage{mathtools}
\usepackage{bm}
\usepackage{mathrsfs}
\usepackage{algpseudocode}
\usepackage[linesnumbered, noend, ruled]{algorithm2e}
\usepackage{hyperref}

\usepackage{blindtext}
\usepackage{graphicx}
\usepackage{subfiles}
\usepackage{enumitem}
\renewcommand{\baselinestretch}{1.0}
\usepackage[margin=0.61in,footskip=0.25in]{geometry}
\def\BibTeX{{\rm B\kern-.05em{\sc i\kern-.025em b}\kern-.08em
    T\kern-.1667em\lower.7ex\hbox{E}\kern-.125emX}}
\begin{document}
\title{Boosting the Convergence of Reinforcement Learning-based Auto-pruning Using Historical Data \\
}

\author{\IEEEauthorblockN{
Jiandong Mu$^1$, Mengdi Wang$^2$, Feiwen Zhu$^2$, Jun Yang$^2$, Wei Lin$^2$, Wei Zhang$^3$
}
\IEEEauthorblockA{
$^{1,3}$Hong Kong University of Science and Technology (HKUST), China\\
$^2$Alibaba Group, China\\
$^1$jmu@connect.ust.hk, $^2${didou.wmd, feiwen.zfw, muzhuo.yj, weilin.lw}@alibaba-inc.com, $^3$wei.zhang@ust.hk}
}

\newcommand{\mengdi}[1]{\textcolor{red}{{\bf Mengdi:} #1}}
\newcommand{\mengdiModify}[1]{\textcolor{orange}{ #1}}

\maketitle

\begin{abstract}
Recently, neural network compression schemes like channel pruning have been widely used to reduce the model size and computational complexity of deep neural network (DNN) for applications in power-constrained scenarios such as embedded systems. Reinforcement learning (RL)-based auto-pruning has been further proposed to automate the DNN pruning process to avoid expensive hand-crafted work. However, the RL-based pruner involves a time-consuming training process and the high expense of each sample further exacerbates this problem. These impediments have greatly restricted the real-world application of RL-based auto-pruning. Thus, in this paper, we propose an efficient auto-pruning framework which solves this problem by taking advantage of the historical data from the previous auto-pruning process. In our framework, we first boost the convergence of the RL-pruner by transfer learning. Then, an augmented transfer learning scheme is proposed to further speed up the training process by improving the transferability. Finally, an assistant learning process is proposed to improve the sample efficiency of the RL agent. The experiments have shown that our framework can accelerate the auto-pruning process by $1.5 \sim 2.5 \times$ for ResNet20, and $1.81 \sim 2.375 \times$ for other neural networks like ResNet56, ResNet18, and MobileNet v1.
\end{abstract}


\section{Introduction}



Nowadays, deep neural network (DNN) has become one of the most popular algorithms for its impressive performance in applications ranging from object detection and image classification to speech recognition. However, the high performance is at the expense of the large model size and huge computing complexity which have blocked DNN from broader usage. As one of the most successful solutions, network pruning \cite{han2015learning}, has been proposed to slim DNN models to obtain a good tradeoff between accuracy and model size, making them feasible for power-hungry devices such as mobile phones. A variety of methods \cite{he2017channel, zhuang2018discrimination, luo2017entropy,  louizos2017learning, tan2020pcnn} have been proposed to prune DNN with different granularities and metrics. 
Channel pruning, which prunes the featuremap and weights in channel granularity, is widely used due to its high efficiency in hardware implementation. In this work, we mainly focused on channel pruning, however, our framework can be easily extended to other pruning schemes.

In order to avoid the extra hand-crafted work introduced by the pruning process and to explore a larger network pruning space, auto-pruning is proposed to compress the network by automatically generating the optimal pruning policy, \emph{i.e.} pruning/preservation ratio of each layer of the input DNN model, using a trainable agent. During the auto-pruning process, the agent is trained by the existing pruning policy and corresponding accuracy. With sufficient data and training time, the agent can converge and generate the optimal pruning policy for the input network. 

Among all the auto-pruning agents, the reinforcement learning (RL) \cite{li2017deep} based agent \cite{he2018amc} has attracted great attention from researchers and developers due to its mature theoretical study, universality, and high performance. However, the training process of the RL-based pruner is time-consuming for several reasons. 
Firstly, the sample efficiency is low since the RL agent is randomly initiated and may not exploit its knowledge of the environment to improve its performance until enough interaction data with the environment are collected. 
Secondly, the computation time of each sample is high as it requires inferences for thousands of images to measure the accuracy. 
The expensive time cost has greatly restricted the usage of RL-based auto-pruning. For example, it takes over one day to prune the ResNet18 automatically on four Nvidia 1080Ti. It will take much longer time for more complex and practical networks like ResNet50, which makes auto-pruning prohibitive for industry, where time-to-market is critical. 

In this work, we propose a comprehensive learning framework to boost the convergence of the RL agent using historical data. 
Firstly, we resort to transfer learning to resolve the random initialization problem. Transfer learning between different pruning ratios, DNN models, and datasets is investigated in detail to accelerate the auto pruning process for the first time. 
Furthermore, we realize that one obstacle to transfer learning between different pruning scenarios is the differences in the preservation ratio, network model,  \emph{etc.} Therefore, we propose network and data augmentation to enhance the transferability so that the converging time of the RL agent can be further reduced compared with vanilla transfer learning. 
Finally, we propose a novel assistant learning process to improve the data efficiency of the RL agent at the beginning of the learning process by generating the training samples according to the pruning history. In this way, the negative effects of the low performance in the initial training process can be minimized.

In summary, we make the following contributions:
\begin{itemize}
    \item We propose to speed up the auto-pruning process with transfer learning. Transfer learning across different pruning ratios, models, and datasets is discussed in a comprehensive manner. 
    \item We propose a novel augmented transfer learning scheme to enhance the transferability between different pruning scenarios, thereby, further reducing the training time.
    \item We propose a novel assistant learning process to improve the data efficiency of the RL agent in the initial training stage. 
    \item Comprehensive experiments have been conducted for the proposed pruning framework. The experiments have shown that the auto-pruning time can be reduced by by $1.5 \sim 2.5 \times$ for ResNet20, and $1.81 \sim 2.375 \times$ for other networks like ResNet56, ResNet18, and MobileNet v1.
\end{itemize}

The remaining paper is organized as follows: Section \ref{background} states the background and related works. Section \ref{framework} introduces the overall framework of this work. Section \ref{tl} and \ref{al} present the proposed augmented transfer learning and assistant learning schemes, respectively, to find the optimal pruning policy. Section \ref{experiments} shows the experiments to validate our proposed learning framework, and Section \ref{conclusion} concludes the paper.

\section{Background and Related Works}
\label{background}

\subsection{Reinforcement Learning-based Auto-Pruning}
There have been a significant amount of works on neural network compression to slim DNNs so that they can be computed efficiently without losing much accuracy. As one of the most important pruning methods, channel pruning reduces the computing complexity by removing the redundant channels on the featuremap. Many channel pruning schemes \cite{he2017channel, zhuang2018discrimination, liu2019metapruning} have been proposed to identify the redundant channels efficiently. 
In our work, the auto channel pruning is achieved by predicting the numbers of the channels preserved in each layer based on the RL agent. Then, the redundant channels are identified and pruned by minimizing the reconstruction error \cite{he2017channel}. An $l_1$ regularization is applied to push the weights of the abandoned channels to zero, and an iterative algorithm is proposed to solve the corresponding Lasso problem \cite{tibshirani1996regression} efficiently. 


RL-based auto-pruning was firstly proposed in \cite{he2018amc}. In that work, a deep deterministic policy gradient (DDPG) agent \cite{lillicrap2015continuous}, which is one of the most popular RL algorithms that target the continuous action space, is utilized to automatically generate the action of each layer of the DNN model. Here the action refers to the preservation ratio, however, its pruning ratio counterpart can be processed in a similar way. Currently, the RL agent takes a long time to converge since it usually takes hundreds of trials to train the DDPG agent, and the accuracy measurement, which requires inferences of thousands of images, is necessary for each trial. Therefore, it is of great importance to improve the convergence speed of the RL agent. 

Several works have proposed to address this problem. In \cite{nair2020accelerating}, previously collected offline data is employed to aid the online learning process by constraining the current policy to stay close to the policy in the previous data. However, this work relies on the assumption that the offline data has the same distribution as the online data, which may not hold true for auto-pruning across different pruning scenarios. Therefore, it cannot be applied in our case. 
In \cite{pertsch2020accelerating}, Pertsch \emph{et al.} accelerates the RL agent by integrating the actions into skills and learns a prior over the skills from the offline data. However, this work also suffers from the offline data consistency problem. 
In our work, we show that our proposed framework can solve this problem by bridging the source task and target task via network augmentation and data augmentation. As a result, the auto-pruning process can be accelerated with historical data from other pruning scenarios. 

\subsection{Transfer Learning}
The core idea of transfer learning, which was proposed in \cite{torrey2010transfer}, is that experience gained in learning of performing one task can benefit the learning performance in related but different task. By transferring the knowledge from a source task to a target task, instead of learning from scratch, the training time can be significantly reduced. As a result, transfer learning has been widely used to expedite the learning process of DNN models. 

Recent works also use transfer learning in RL \cite{taylor2009transfer, zhu2020transfer} to accelerate the learning process by leveraging and transferring external expertise. However, there still lacks investigations of transfer learning in auto-pruning to the best of our knowledge. 
Furthermore, vanilla transfer learning may suffer from significant performance degradation in several scenarios and become less valuable for practical usage due to the difference between the source task and the target task. 
In this work, we present a framework to speed up the RL-based auto-pruning via transfer learning for the first time. Then we further boost the performance of the transfer learning by network and data augmentation. 

\section{Framework}
\label{framework}
\begin{figure}
    \centering
    \includegraphics[width=1\columnwidth, clip=true]{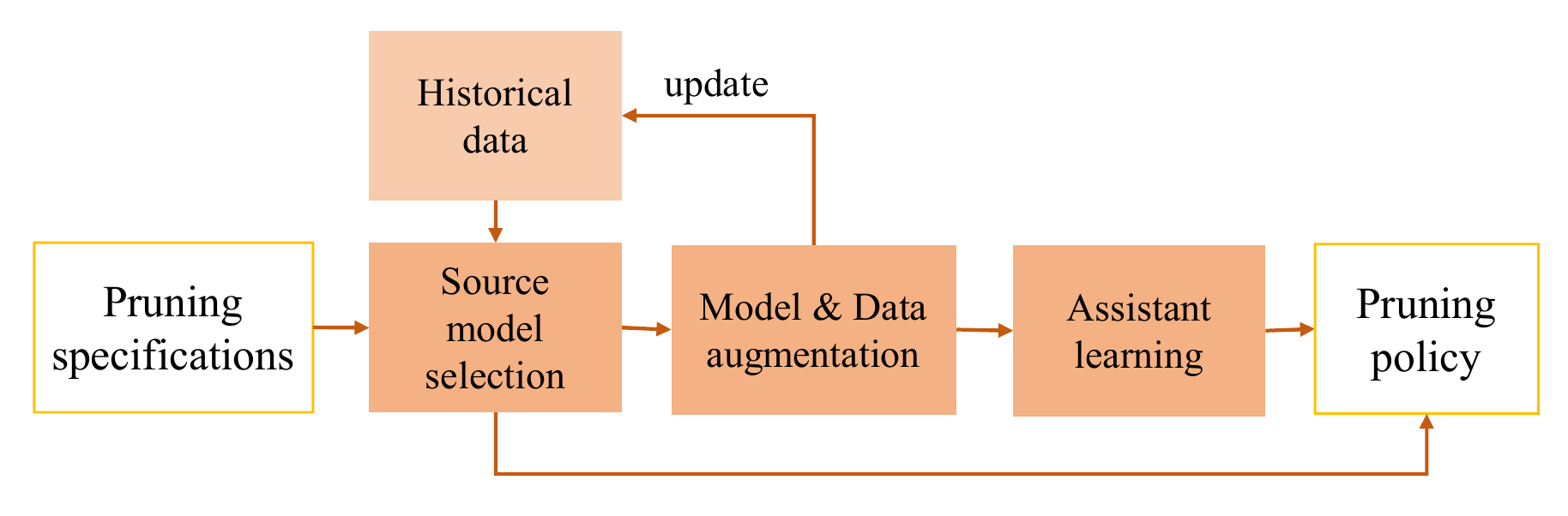}
    \caption{Proposed historical data based framework to boost the convergence of RL agent in auto-pruning.}
    \label{fig:framework}
\end{figure}

In this section, we will show the overview of the proposed framework as illustrated in Fig.\ref{fig:framework}, followed by a detailed introduction of each component. 

In this work, we aim to accelerate the RL-based auto-pruning process by taking advantage of the historical data in other pruning scenarios. 
To achieve this goal, the pruning specifications (\emph{e.g.} DNN to be compressed, preservation ratio \emph{etc.}) and historical data are first imported to the source model selection part. The source model can be selected according to the experience and the performance of the vanilla transfer learning between historical model and target model. 
If the time constraints are already satisfied, the RL agent can be directly accelerated by vanilla transfer learning. Otherwise, an augmented transfer learning process will be conducted to boost the transferability of the source model so that the converging time of the transfer learning can be further reduced. 

Our proposed augmented transfer learning consists of model augmentation and data augmentation, which increase the transfer learning efficiency by augmenting the model of the actor neural network and the data in the replay buffer, respectively. The corresponding historical model will also be updated to ease the later transfer learning. 

Finally, assistant learning will be employed to improve the sampling efficiency in the initial phase of the RL process. This is based on the rationale that the RL agent suffers from low sampling efficiency since the interactions with the environment in the initial period are random. We can boost the convergence of the RL algorithm by taking advantage of the historical data from other pruning scenarios to improve the sampling efficiency in the early RL phase.

\section{Augmented Transfer Learning}
\label{tl}
Similar with previous works \cite{he2018amc, wu2018pocketflow}, we consider the auto channel pruning as a Markov process \cite{gagniuc2017markov} in which channels are pruned in a layer by layer manner. The layer information \emph{e.g.}, the layer index, height, width \emph{etc.}, is considered as the state, while the preservation ratio is considered as the action. 
The DDPG-based RL agent \cite{lillicrap2015continuous}, which consists of an actor and a critic, is employed to predict the preservation ratio of each layer of the input neural network automatically. The actor, which is usually a light-weight CNN model, exploits the information of the states and generates the action. The critic, which is also a light-weight CNN model, is used to evaluate the action to suggest the optimizing direction of the actor. During the training process of RL agent, the weights of the actor network and critic network are updated by maximizing the output of the critic network and balancing the Bellman equation \cite{sutton2018reinforcement}, respectively. As the actor and critic converge, the optimal pruned model can be obtained by measuring the preservation ratio of each layer according to the inference of the actor network given the input state information. 

However, the actor network and critic network take a long time to converge due to the huge number of the parameters in their networks and randomness of the network initialization. 
Observing that the training process of a DNN can be accelerated by transfer learning, which reuses the weights of the source task as the starting point of the target task, 
we find transfer learning promising to boost the convergence of RL-based auto-pruning, since there are many common features between auto-pruning agents in different scenarios. For example, it has been observed that the optimal pruning policies for different preservation ratios have a similar pattern, which depends on the importance of each layer. Hence, we proposed to embed the transfer learning framework into the RL-based auto-pruner so that the weights trained in other scenarios can be reused to speedup the convergence of the current pruning agent. 

\begin{figure}
    \centering
    \includegraphics[width=1\columnwidth, clip=true]{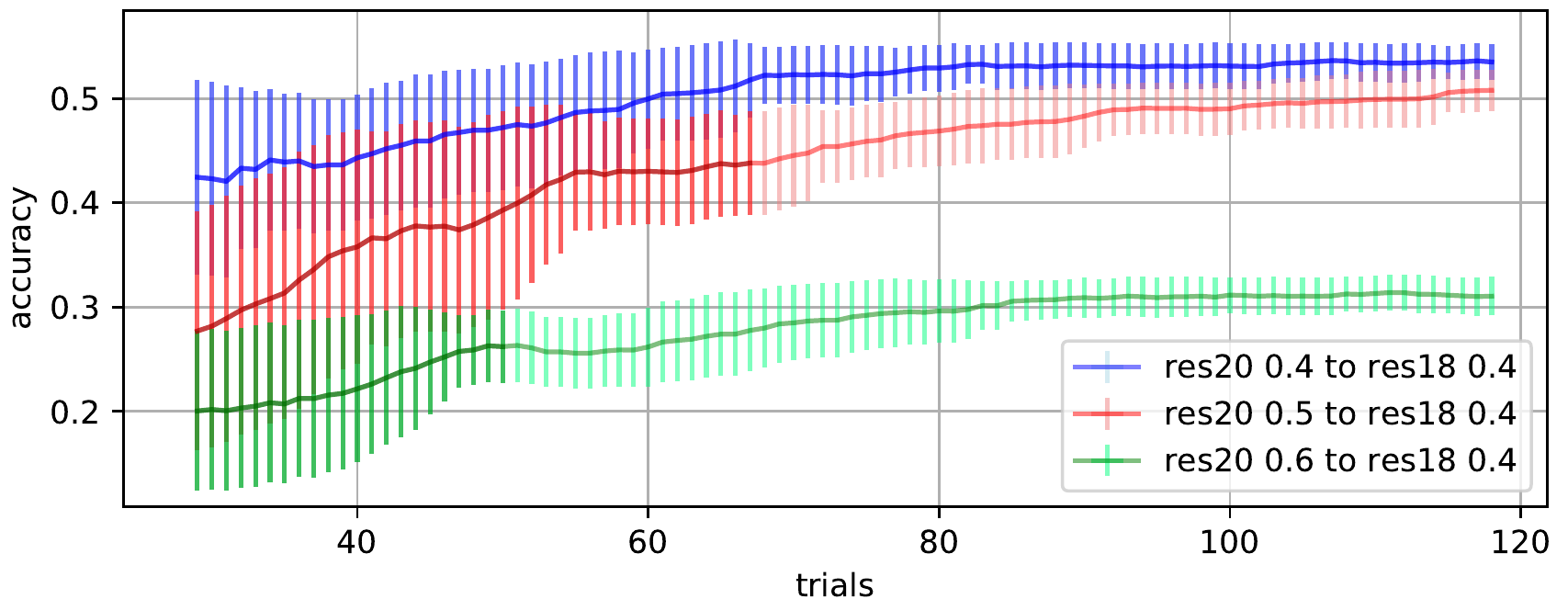}
    \caption{Transfer learning from different starting points. \textit{"res20 0.4 to res18 0.4"} refers to transfer learning from ResNet20 with a preservation ratio of 0.4 to ResNet18 with a target preservation ratio of 0.4. The rest of the legend can be interpreted in the same manner.}
    \label{fig:ckpt_selection}
\end{figure}

\begin{figure}
    \centering
    \includegraphics[width=1\columnwidth, clip=true]{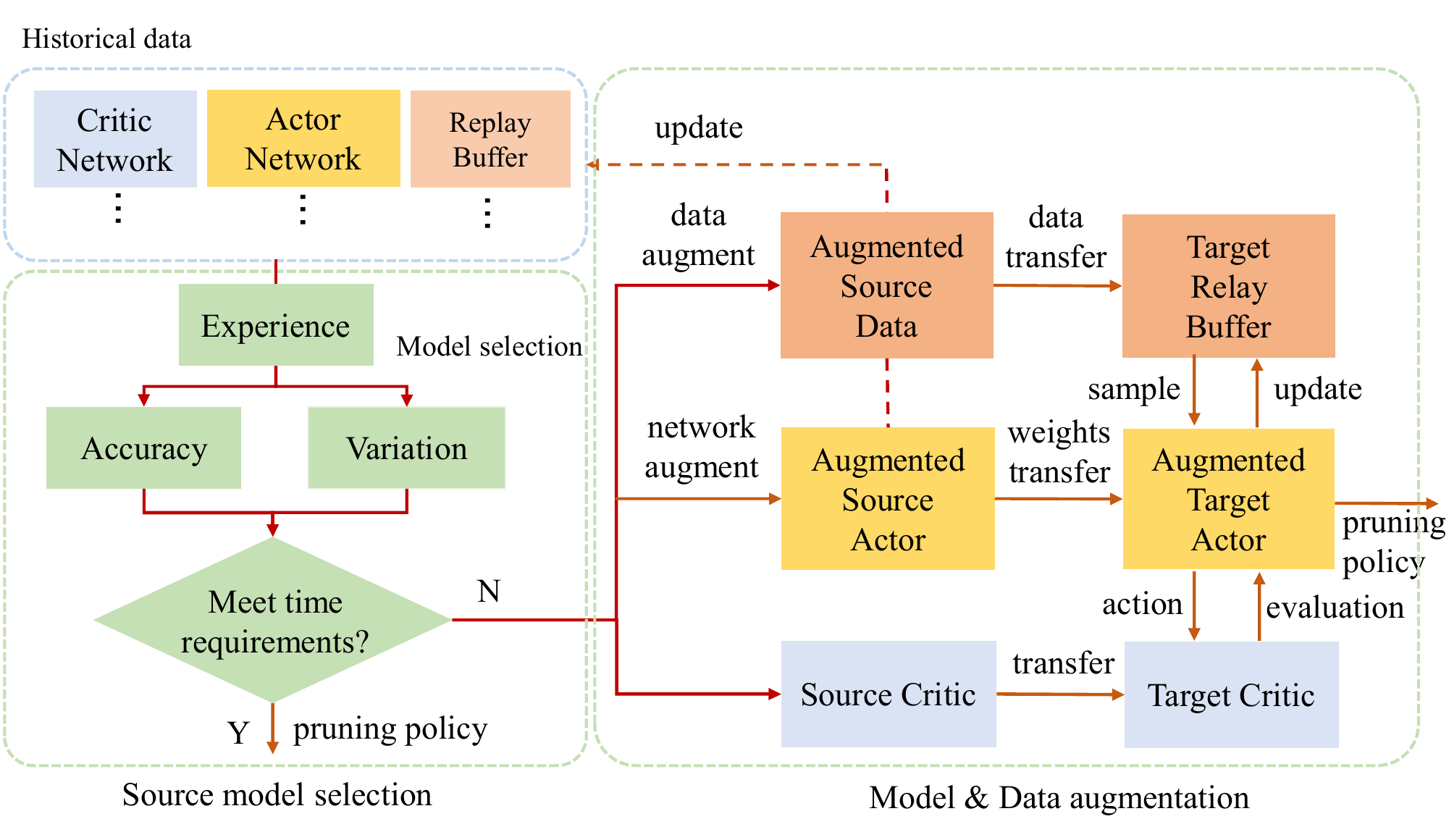} 
    \caption{Our proposed augmented transfer learning. Source selection, network augmentation, and data augmentation are applied to boost the convergence of the transfer learning process.}
    \label{fig:tl_framework}
\end{figure}

Vanilla transfer learning may not work perfectly across different pruning ratios, models, and datasets. For example, the experiments for transfer learning from ResNet20 to ResNet18 are shown in Fig. \ref{fig:ckpt_selection}. It is obvious that transfer learning from ResNet20 with preservation ratio of 0.5 and 0.6 have non-optimial performance. 
This is caused by the inconsistency of the source preservation ratio and the target preservation ratio. The high preservation ratio of the source model may lead to high action in the target model. However, when the weights of the actor are transferred to a pruning scenario which has a lower preservation ratio, it tends to mislead the target model, and the predicted action for the initial layers becomes higher than expected. Given the constraints of the overall preservation ratio requirement, the action for the later layers will be suppressed and become lower than expected. This inconsistency will significantly harm the performance of the RL-based auto-pruning agent. 
In order to solve this problem, we propose augmented transfer learning, which consists of an automatic source selection scheme, network augmentation, and data augmentation, to transfer the knowledge between different pruning scenarios efficiently. The framework of the augmented transfer learning is illustrated in Fig. \ref{fig:tl_framework}.

In our augmented transfer learning framework, we first select the source model candidates according to the experience. For example, transfer learning from high preservation ratio to low preservation usually indicates a less satisfying result, as shown in \ref{fig:ckpt_selection}, therefore transfer learning from low preservation ratio is usually preferred. Then the source models can be further filtered by transfer learning from multiple starting points in a parallel manner and the optimal source model can be selected according to our source selection scheme, which will be explained in detail in \ref{source_selection}. 
Then, a decision is made according to the convergence speed of the vanilla transfer learning. If the convergence speed meets the time requirement specified by the user, the auto-pruning policy can be obtained by the inference of the source actor network. Otherwise, data and network augmentation will be applied to enhance the transferability. 
Next, the augmented data and network will be transferred to the target buffer and augmented target actor, respectively. The actor will be trained according to the samples from the replay buffer and the optimization direction given by the critic network. As the augmented target actor converges, the pruning policy can be obtained by the inference of it. 
Finally, the augmented data and actor will be used to update the library to ease the later auto-pruning process.

The source selection, network augmentation, and data augmentation are explained in the following subsections. 

\subsection{Source Selection}
\label{source_selection}
As Fig. \ref{fig:ckpt_selection} shows, it is critical to select the starting point of the target network. A good source network may benefit the training process, while an inappropriate source network may poison the performance. However, it will be time-consuming to finish the transfer learning process for all of the source models in the library. 
Hence, we exploit an early stopping scheme to search for the optimal source model efficiently. 

In this scheme, transfer learning from multiple pre-trained models starts at the same time, and a moving average-based smoothing process, which has a window size of $21$ based on our experiments, is conducted for each learning curve, as shown in Fig. \ref{fig:ckpt_selection}. The mean of the points inside the window is considered as the value of the center points, while the variance of the points inside the window is a good approximation of the corresponding variance. A minimum of $30$ trials are required to train each transfer learning process due to the high randomness at the beginning of the training process. Then the optimal transfer learning starting point can be selected according to the priority of different source models, which is defined as the following. 

\textit{Def: We consider source model "A" as significantly superior to source model "B" if there exists an trial $x$ in which $avg_A(x) - avg_B(x) > var_A(x) + var_B(x)$. }

Thus, the transfer learning process from the non-optimal source model can be stopped early to save resources, as illustrated by the green line and red line in Fig. \ref{fig:ckpt_selection}. 
Note that there exists the case that one model may not be significantly superior to the other if the performance difference of the two source models is smaller than the corresponding variance. In this case, we choose the source model that has the better inference accuracy at the maximum trial specified by the user.

\subsection{Network Augmentation}
The actor network and critic network inside the RL agent aim to predict the action for the given input states and to evaluate its corresponding performance. However, the transferability is not considered in the actor network design, which may lead to sub-optimal transfer performance. We show that by modifying the network, the transferability can be significantly increased without causing extra computational costs. 

To achieve this goal, we modify the actor network to generate invariant defined in Eqn. \ref{invariant} across different scenarios instead of the pruning ratio of the given layer, which may vary significantly for different pruning ratios and models. As a result, the weights of the actor network for different pruning scenarios can be easily reused with little re-training process. 
We consider the following invariant $I_k$ between different pruning scenarios $c$ for layer $k$: 
\begin{equation}
    I_k = \frac{a_{k}^{c}}{p^{c}}
    \label{invariant}
\end{equation}
where $a_{k}$ indicates the action, \emph{i.e.} the preservation ratio of layer $k$, and $p$ is the targeted preservation ratio. 
We approximate the invariant using the actor network inside the RL agent with $I_k = f(s_k)$, where $s_k$ indicates the input states for layer $k$. 
This is based on the assumption that the preservation ratio of each layer $a_{k}^{c}$ for pruning scenario $c$ is proportional to the overall preservation ratio $p^{c}$. Therefore, $\frac{a_{k}^{c}}{p^{c}}$ is invariant to different pruning scenarios and can be easily transferred. 
We note that this assumption may not hold true for transfer learning across different models since the invariant depends on $k$. However, it turns out to be a good assumption to bridge pruning scenarios with different preservation ratios, and the experiments have validated this approximation. 

By predicting the above invariant using the actor network, the actual action ratio can be obtained by $f(s_k)*p^c$, which has little computing overhead. 

\subsection{Data Augmentation}
\label{data_aug_subsection}
Another problem of the RL agent comes from the buffer filling process at the beginning of the training stage. Note that the adjacent states are strongly dependent. Therefore, a replay buffer is employed to store and shuffle the training samples, \emph{e.g.}, states information, actions, and corresponding rewards, to resolve the dependence. Though the replay buffer can achieve decent performance improvement, the filling process is time-consuming and exacerbates the training costs due to the high expense of each sample. However, on the other hand, this also opens up another opportunity for accelerating the RL-based auto-pruning. 
To solve this problem, we propose to transfer training data to provide more information between the source model and the target model, in addition to weight-based transfer learning between the models inside the RL agents. As a result, the transfer efficiency can be further improved. However, directly applying the training data of the source pruning agent may cause bias of the target agent, since the pruning scheme of the source agent and target agent may be different. In this work, we propose the following data augmentation to reduce the data bias between the source and target models. 




Firstly, we consider data augmentation for transfer learning across the same model with different preservation ratios. The rationale is that a high preservation ratio usually indicates a high action for each layer. To reduce the action bias caused by the inconsistency of the preservation ratio for the source task and target task, an action scale is introduced to fine-tune the action. 
In this work, we scale the action by the following formula, which aims to protect the critical layers from pruning: 
\begin{equation}
    a_k^{target} = 1 - \frac{1-a^{source}_k}{1-p^{source}} (1-p^{target}). 
    \label{data_aug_equation}
\end{equation}
For example, when layer $k$ is critical and $a_k^{source}$ is close to $1$, the second term will be close to zero. As a result, $a_k^{target}$ will be close to $1$ as well.

Next, we consider the data augmentation for transfer learning across different models and datasets. In this case, the samples are fine-tuned according to the prior information of the source and target model. We illustrate this by the widely used ResNet model. It can be observed that the shortcut layer and the top layer inside the ResNet block have fewer parameters than other layers. As a result, the preservation ratios for these layers are usually high since pruning these layers may lead to a significant accuracy loss. Therefore, it becomes natural to adjust the preservation ratio according to the importance of the layers for the transfer learning between different ResNet models. 
In our experiments, we set the preservation ratio to $1$ for the layers that are critical in the target model. The preservation ratios for layers that are critical in source model, while not critical in the target model, are uniformly reduced to maintain the overall preservation ratio unchanged. Then, Eqn. \ref{data_aug_equation} can be used to reduce the action bias if the inconsistency of the preservation ratio exists. 

Finally, these data are randomly sampled to train the auto-pruner which is similar to the data sampling in the vanilla transfer learning. 
The experiments presented in Section \ref{experiments} will show the effectiveness of our proposed augmentation schemes.


\section{Assistant Learning}
\label{al}
In the previous section, we took advantage of the previous well-trained model and data to improve the learning speed of the target model. 
However, the learning time is still intolerable in many time-critical scenarios. This is caused by the low sampling efficiency of the RL algorithm, since the actor network is not well-tuned and the action generated by the actor network in the initial trials may be useless.  

Inspired by the high data efficiency of Bayesian optimization, where each sample is generated according to the historical samples and a well-defined acquisition function \cite{frazier2018tutorial}, we propose an assistant learning process to generate the next training samples at the initial stage according to the training history. As a result, the sub-optimal samples generated by the actor network at the initial stages can be avoided and the corresponding data efficiency can be significantly improved. 
After a certain number of trials, ($30$ in our experiments), the RL actor network becomes mature and the transfer learning based RL algorithm, which trains the network according to the action generated by the actor, can be resumed. 

\begin{figure}
    \centering
    \includegraphics[width=1\columnwidth, clip=true]{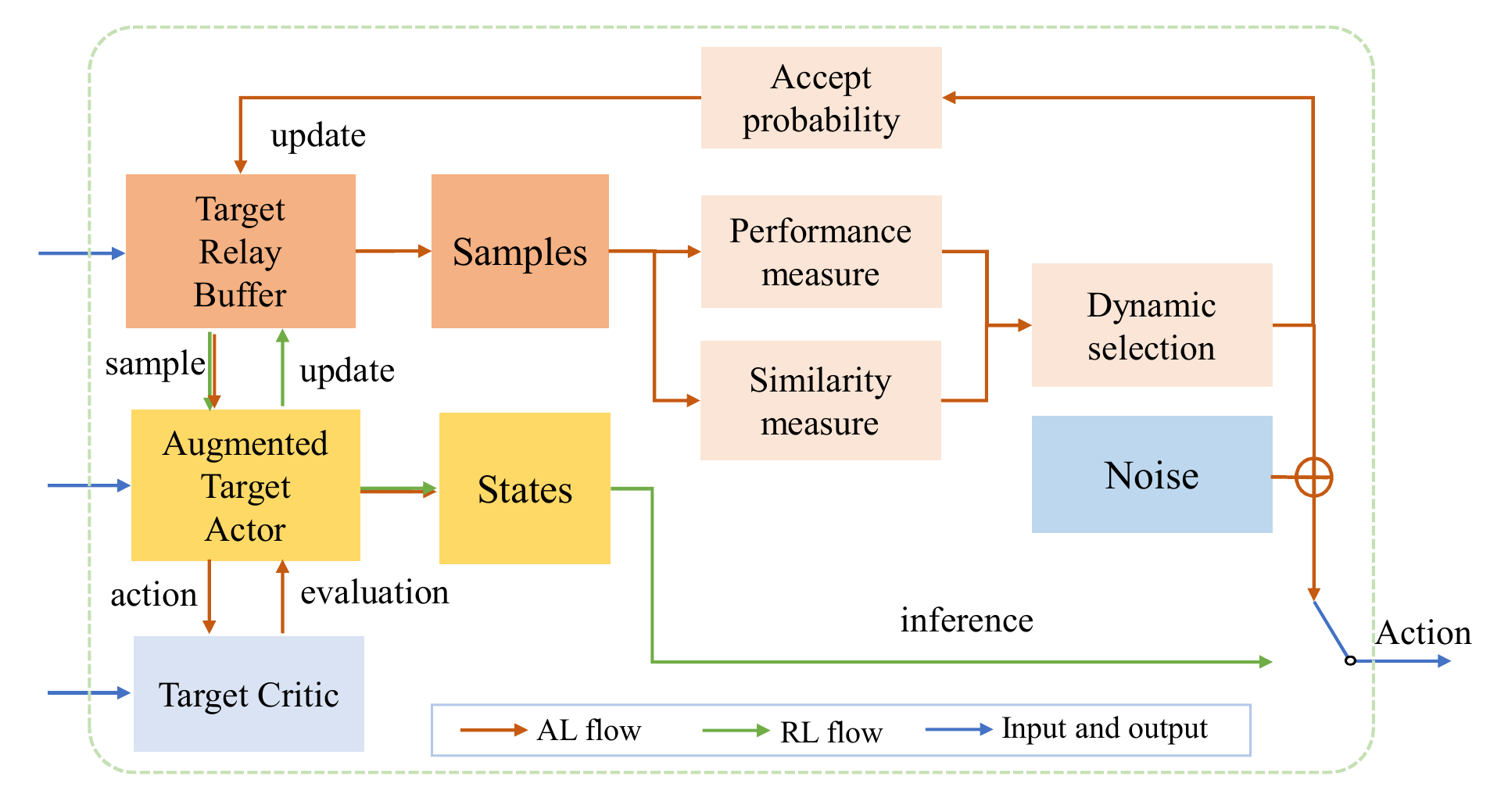}
    \caption{Our proposed assistant learning. Inspired by the Bayesian Optimization, the historical data are employed by a measuring and a dynamic selection process to improve the sample efficiency at the initial learning stages.}
    \label{fig:al_framework}
\end{figure}

The details of our proposed assistant learning algorithm are shown in Fig. \ref{fig:al_framework}. 
The proposed algorithm contains two separate dataflows, as indicated by red and green flow, respectively, in Fig. \ref{fig:al_framework}. The green color refers to the RL-based auto-pruning process, which gets the action by the inference of the actor neural network. The dataflow highlighted in red represents the history data based action prediction flow to improve the data efficiency of the RL agent at initial stages. A switch is used to select the action from the two data paths and output the action to interact with the environment, which is network pruning in our case. The predicted action will also be used to update the target replay buffer with an accept probability, which we will explain later.

It is non-trivial to predict a sample according to the states and actions of the historical pruning data due to the large pruning space and numerous historical records. In order to exploit the pruning history efficiently, the samples of the pruning history are iterated, and the scores for similarity and performance are evaluated for each historical record. Historical data with higher similarity and performance are preferred for later action generation. This is based on the assumption that the design space of the pruning network is continuous and the samples with similar input state information should have a similar action. This is a rational assumption since the accuracy changes continuously as the pruning ratio and the convolution size change. 

The definition of the similarity $S$ can be illustrated by the following formula:
\begin{equation}
    S = \prod_x G \left( h(x), i(x), \sigma \right),
\end{equation}
where $G$ refers to the Gaussian function to indicate the distance between historical states $h(x)$ and input states $i(x)$, $x$ is the iterator of the states, and $\sigma$ is the standard deviation of $G$ to adjust the distance measurements. In this work, we set $\sigma$ to $0.1$ according to experience. 
For the performance measurement of the historical data, the inference accuracy of the corresponding pruned network is considered as a good indicator. Metrics like power, throughput, \emph{etc.} can also be considered for pruning with hardware constraints, which is temporarily not discussed in this work.

Then the historical data can be selected according to both the similarity and the performance metric, as shown in the following formula:
\begin{equation}
    M = S^{\omega} + P,
\end{equation}
where $M$ is the final selection metric, $S$ refers to the similarity measurements, $P$ is the performance measurements, and $\omega$ is a weight parameter. In our experiment, we set $\omega = 2$ according to the experimental performance. 
Our dynamic selection scheme selects historical samples based on the metric $M$ to cater to different pruning scenarios. In the early trials, exploration is more important. Therefore, the top $n$ ($n=3$ in our experiments) historical samples  are randomly selected to explore wider policy spaces. In the later trials, accuracy becomes more critical. The historical sample with the highest $M$ is selected to boost the performance of the RL agent.

Then, the action of the input state can be given according to the selected historical record and noise, which also aims to widen the exploration space of the RL process. Note that it is critical to select appropriate noise to get a good tradeoff between exploration and exploitation. Both uniform noise and Gaussian noise are investigated in our experiments, and we found that uniform noise is more effective for the assistant learning process. Then the uniform noise passes through a linear decay filter to reduce the disturbance as the actor network converges.  

Finally, the new interactions with the environment are feedback to the replay buffer to update the data in the buffer. In contrast with the RL agent, which directly accepts all the data and updates the buffer, we adopt a probability-based updating algorithm to get rid of the low-performance trials in the assistant learning phase. This is because the RL agent focuses on the whole training phase, while our assistant learning algorithm only focuses on the initial training phase where low-quality samples are common. Therefore, accepting all the samples will harm the convergence of the actor, and a probability-based accepting rule becomes necessary. In our case, the trials with accuracy ranks in the top $1/3$ of the batch can be accepted and update the buffer without rejection. The other samples will be accepted with an exponential decreased probability.  

\begin{figure}
    \centering
    \includegraphics[width=1\columnwidth, clip=true]{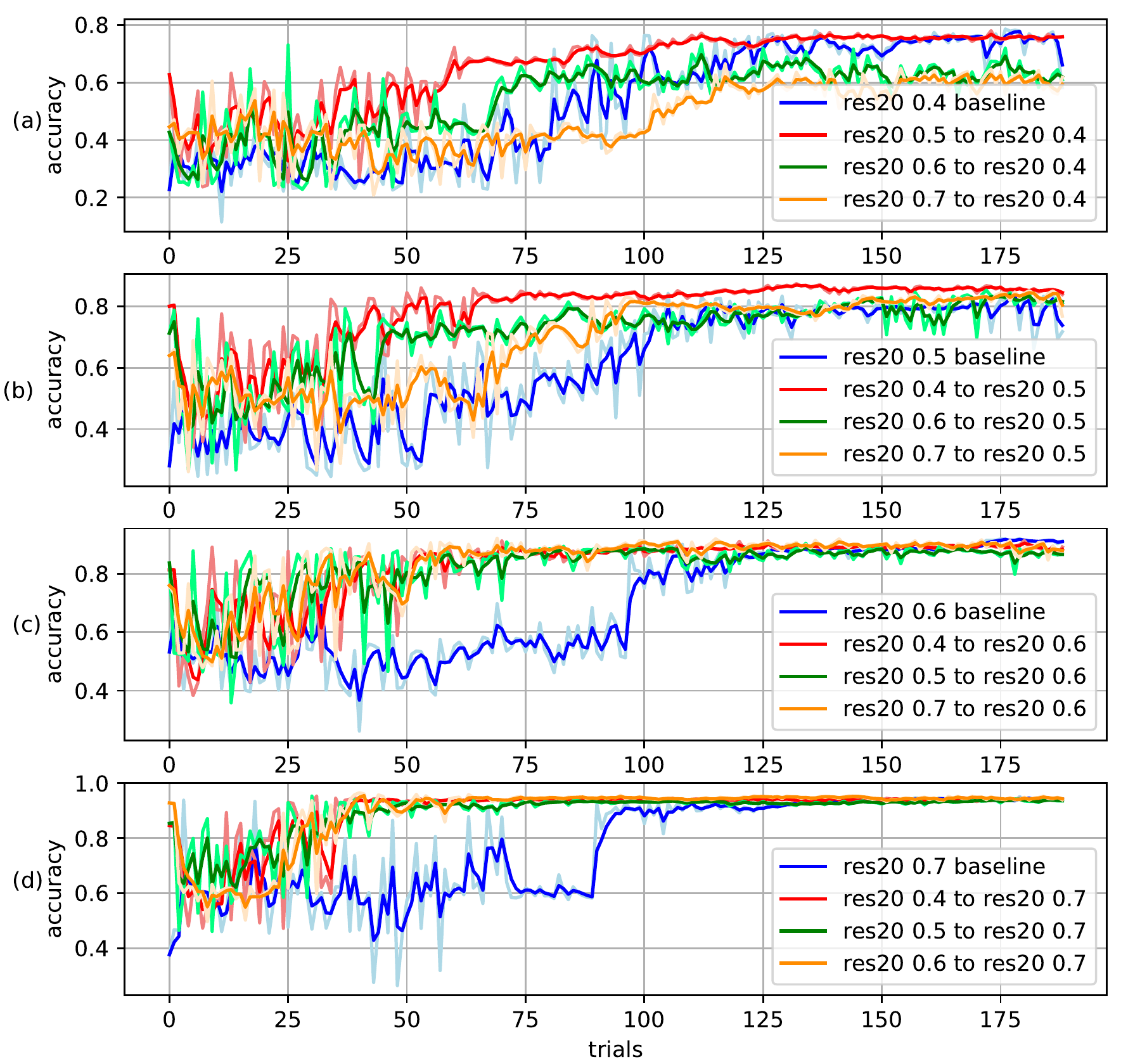} %
    \caption{Experiments for vanilla transfer learning between pruned ResNet20 model with different preservation ratios. The baseline is based on PocketFlow \cite{wu2018pocketflow}}
    \label{fig:transfer_learning_ratios}
\end{figure}

We also investigated model-based action predictors to take advantage of the history knowledge and to predict the action. XGBoost \cite{chen2016xgboost} and Matrix Factorization \cite{koren2009matrix} based models have been tried. Both the simulated annealing based solver and Adam \cite{kingma2014adam}, which is embedded in Tensorflow, have been employed to solve the maximum point of the model. However, we found that this is non-practical due to the huge data requirement to learn the model and the low generalizability of the model across different pruning scenarios. 





\section{Experiments}
\label{experiments}
Our framework has run on an Intel(R) Core(TM) i7-5820K CPU @3.30GHz with a 32 GB DDR memory. Tensorflow 1.12 is employed for the auto-pruning process. The experiments are conducted on Nvidia GeForce GTX TITAN X, which has 3072 cores and a boost frequency of 1089MHz, leading to a peak throughput of 6691 GFLOPS. The GPU cards are connected with the host machine via a PCI-e 3.0 interface which offers a maximum bandwidth of 8GT/s. The widely used CUDA-10.1 is used to program the DNN applications on GPU efficiently.

The vanilla RL agent-based auto-pruner, which is conducted on the auto-pruning platform PocketFlow \cite{wu2018pocketflow}, is employed as the baseline. Note that the accuracy in the following experiments refers to the inference accuracy of the pruned model without the fine-tuning process. The learning curve is smoothed using the exponential moving average, which is also built into the widely used TensorBoard platform, with a weight factor of $0.5$. 




\begin{figure}
    \centering
    \includegraphics[width=1\columnwidth, clip=true]{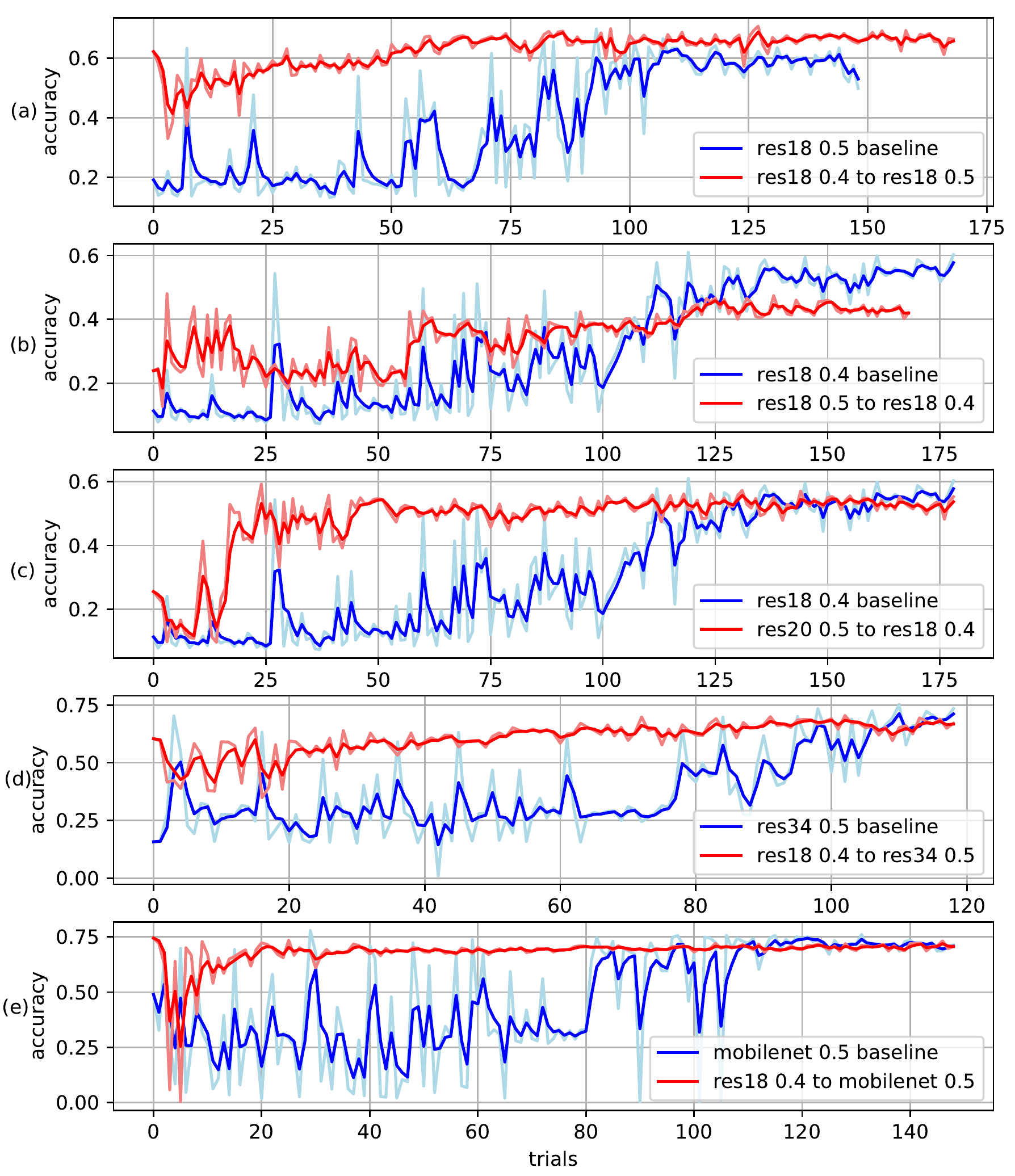} 
    \caption{Experiments for vanilla transfer learning in five pruning scenarios (a)-(e). The baseline is based on PocketFlow \cite{wu2018pocketflow}}
    \label{fig:transfer_learning}
\end{figure}

\subsection{Vanilla Transfer Learning}
In this subsection, comprehensive experiments are conducted to show the performance of the vanilla transfer learning. The results are illustrated in Fig. \ref{fig:transfer_learning_ratios} and Fig. \ref{fig:transfer_learning}. 

Firstly, we show the performance of the transfer learning across different preservation ratios. To have a comprehensive and fair comparison, we provide experiments for all the transfer learning cases within the ratio list (0.4, 0.5, 0.6, 0.7), as illustrated in Fig. \ref{fig:transfer_learning_ratios}. In Fig. \ref{fig:transfer_learning_ratios}(a), we show the RL based auto compression for ResNet20 with a target preservation ratio of $0.4$. Among all the curves, "res20 0.4 baseline" refers to the accuracy of the auto-pruning process in PocketFlow \cite{wu2018pocketflow}, which serves as the baseline in our experiments. "res20 0.5 to res20 0.4" indicates transferring the knowledge learned in the source pruning scenario, which has a preservation ratio of $0.5$, to the target pruning scenario with a preservation ratio of $0.4$. The legend for other figures can be interpreted in a similar manner. 

In this experiment, we made the following observations. (1), Auto-pruning for high preservation ratio is more likely to benefit from the transfer learning process. In (c) and (d), vanilla transfer learning can significantly boost the convergence of the RL algorithm. However, in case (a) and (b), the benefits of transfer learning diminishes. In case "res20 0.7 to res20 0.4", the transfer learning may even poison the convergence of the auto-pruning process. (2), source models with lower preservation ratios are preferred. For example, in (b), "res20 0.4 to res20 0.5" has higher performance than "res20 0.6 to res20 0.5". 
These are because of the inconsistency of the preservation ratios as we mentioned in section \ref{tl}. In precis, although the transfer learning for (c) and (d) are promising, the performance in (a) and (b) are far from optimal, this motivates us to develop the novel framework to boost the vanilla transfer learning process.

\begin{figure}
    \centering
    \includegraphics[width=1\columnwidth, clip=true]{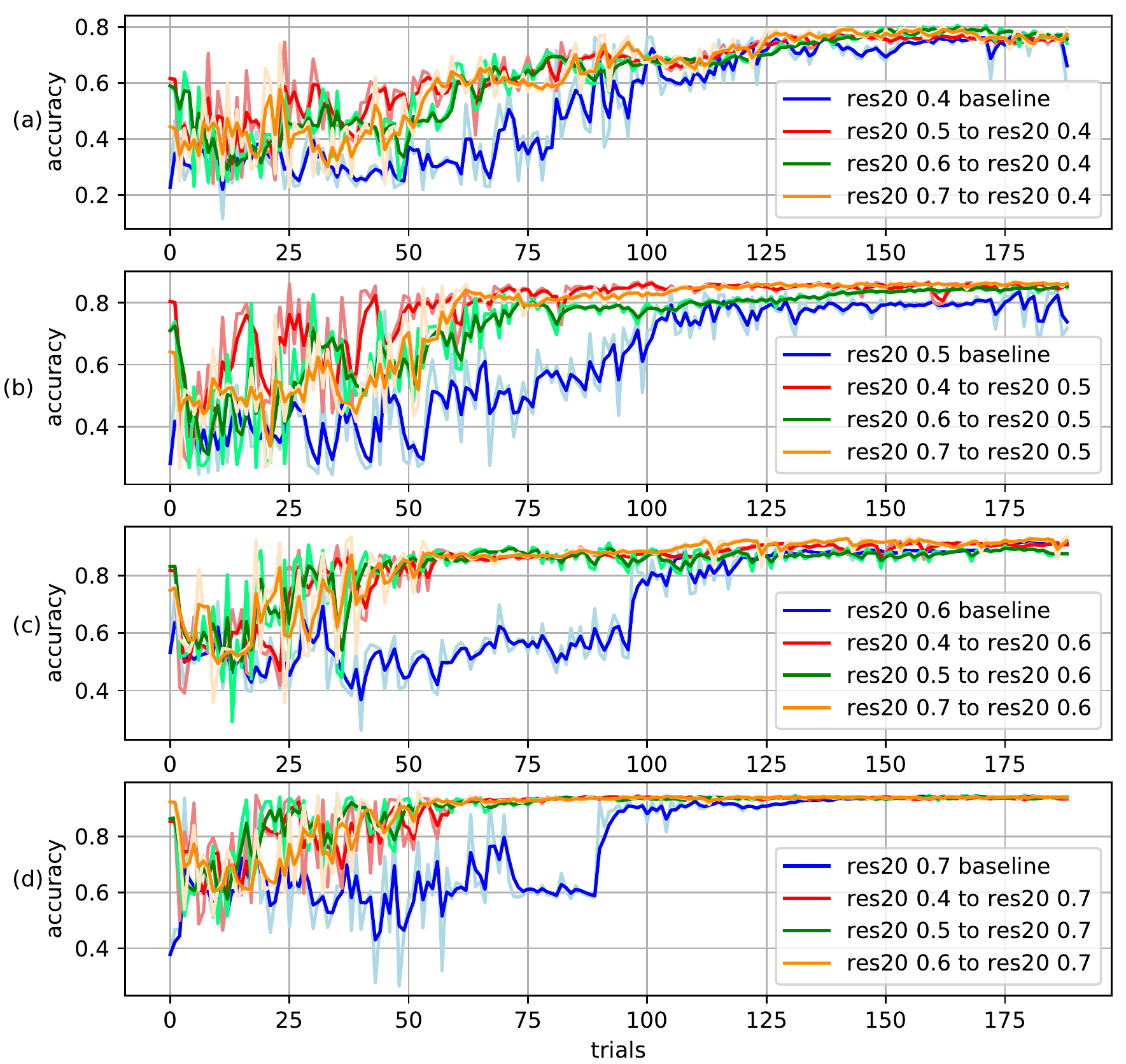}
    \caption{Experiments for augmented transfer learning for ResNet20.}
    \label{fig:nt-aug}
\end{figure}

We further extend our transfer learning algorithm to different datasets and DNN models as shown in Fig. \ref{fig:transfer_learning}. Fig. \ref{fig:transfer_learning} (a), (b) and (c) shows the transfer learning based auto-pruning for ResNet18. 
In Fig. \ref{fig:transfer_learning}(a), the source model has a preservation ratio of $0.4$, while the target model has a preservation ratio of $0.5$. 
It can be observed that transfer learning can significantly increase the convergence speed. The baseline RL agent converges in around $100$ trials, while the transfer learning based RL agent can achieve the same pruning accuracy within $60$ trials, and acceleration of $1.67 \times$ is achieved. The transfer learning process can also benefit the final accuracy of the auto-pruner. 

In Fig. \ref{fig:transfer_learning}(b), we show the transfer learning for ResNet18 with different source and target preservation ratios, \emph{i.e.} "res18 0.5 to res18 0.4". 
In contrast with the previous example, although the transfer learning based auto-pruner in this scenario has a better pruning accuracy at the initial trials, it fails to outperform the baseline when the training process converges. This indicates the fact that vanilla transfer learning may not always benefit the learning process, and may lead to degraded performance. However, in later experiments, we show that this problem can be avoided by our proposed augmented transfer learning and assistant learning. 

\begin{figure}
    \centering
    \includegraphics[width=1\columnwidth, clip=true]{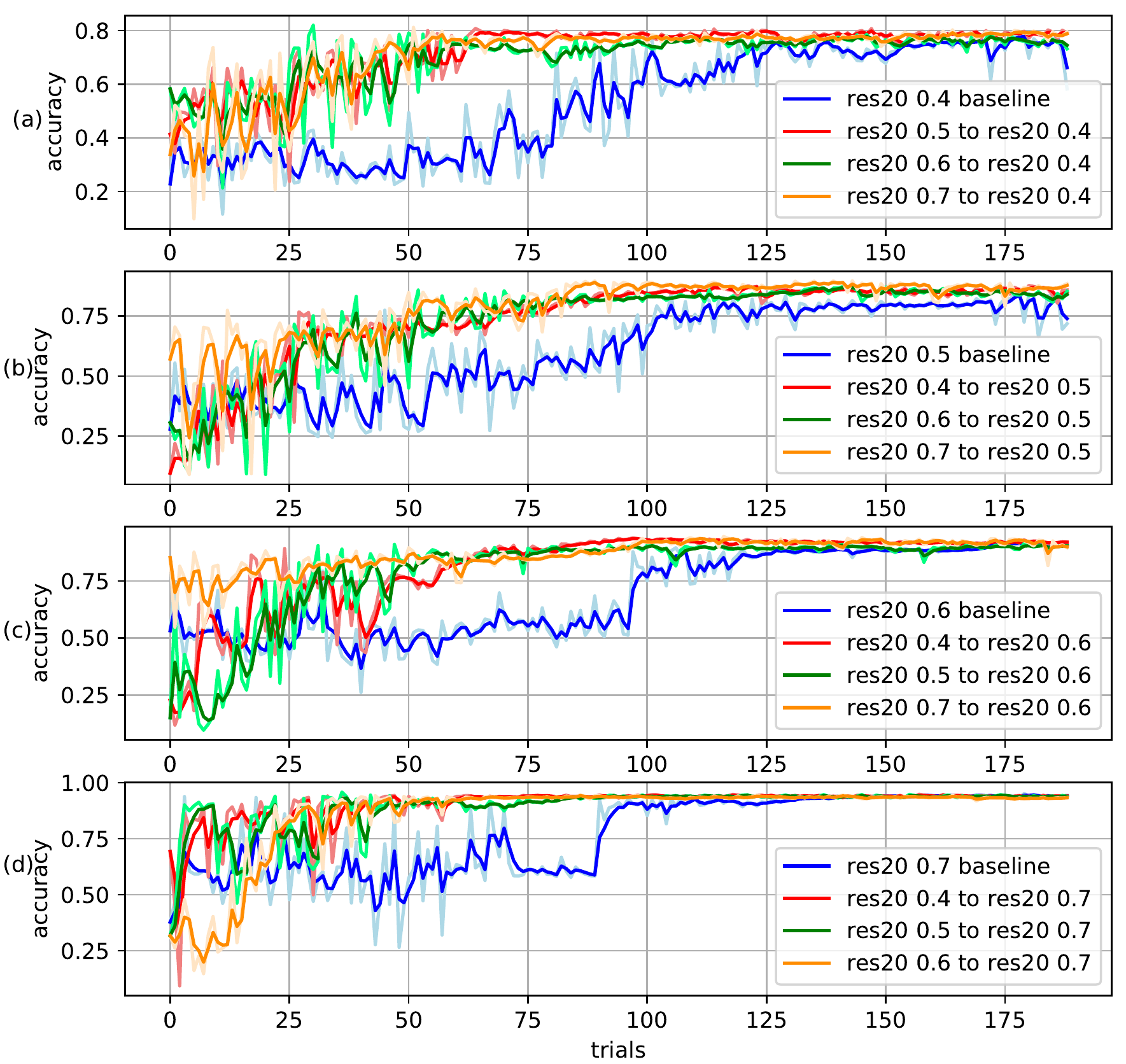}
    \caption{Experiments for assistant learning for ResNet20.}
    \label{fig:al_exp}
\end{figure}

Fig. \ref{fig:transfer_learning}(c) shows transfer learning for ResNet18 from different models and dataset. In this case, the baseline converges in $125$ trials while the transfer learning based counterpart can converge in $25$ trials. Therefore, around $5 \times$ speedup can be achieved.

In Fig. \ref{fig:transfer_learning}(d),(e), we show the transfer learning for other DNN models. Transfer learning from ResNet18 to ResNet34 is shown in Fig. \ref{fig:transfer_learning}(d), 
while transfer learning from ResNet18 to a light-weight network running on embedded systems like MobileNet is shown in Fig. \ref{fig:transfer_learning}(e). 
In both cases, a significant speedup, which is around $5 \times$, in convergence time can be observed.

In summary, our experiments in this section show that the transfer learning process can boost the learning process by $1.67 \sim 5 \times$ for different pruning scenarios. However, it may harm the accuracy in a few cases, for example, Fig. \ref{fig:transfer_learning_ratios}(a) and Fig. \ref{fig:transfer_learning}(b). In the later experiments, we show that our proposed algorithms can solve this performance degradation problem and further speed up the learning process. 

\subsection{Augmented Transfer Learning}
In this section, we propose experiments to verify the validity of our augmented transfer learning algorithm. In order to have a fair comparison, we conducted experiments for all the transfer learning scenarios within the preservation ratio list (0.4, 0.5, 0.6, 0.7), which is similar to Fig. \ref{fig:transfer_learning_ratios}. The corresponding experiments are shown in Fig. \ref{fig:nt-aug}. 
In (c) and (d), we observe that the performance of the augmented transfer learning is similar to its counterpart in Fig. \ref{fig:transfer_learning_ratios}, since they are already close to optimal.
However, in (a) and (b), we found that the converging speedup can be significantly increased. We also observe a substantial improvement in the final accuracy of the pruned model. More specifically, the accuracy loss problem in "res20 0.6 to res20 0.4" and "res20 0.7 to res20 0.4" as we have mentioned in the previous subsection has been solved. 

We summarized the experiments for augmented transfer learning for other models in section \ref{exp_diff_models} to have a more clear comparison.

\begin{table}[t]
    \centering
    \caption{Boosting the convergence of ResNet56 with preservation ratio of 0.4}
    \begin{tabular}{|c|c|c|c|}
         \hline
         pruning scenarios & source model & convergence time (trials) & accuracy \\
         \hline
         baseline \cite{wu2018pocketflow} & None & 100 & 0.8675 \\
         \hline
         vanilla transfer & res20 0.4 & 50 & 0.8479 \\
         \hline 
         augmented transfer & res20 0.4 & 50 & 0.8838 \\ 
         \hline
         assistant Learning & res20 0.4 & \textbf{50} & \textbf{0.9078} \\
         \hline
         vanilla transfer & res20 0.7 & 75 &  0.8638\\ 
         \hline 
         augmented transfer & res20 0.7 & 55 &  0.8624\\ 
         \hline
         assistant learning & res20 0.7 & 55 &  0.8694\\ 
         \hline
    \end{tabular}
    \label{tab:res56}
\end{table}

\subsection{Assistant Learning}
In this subsection, we provide the experiments for auto-pruning after the assistant learning process. Note that the assistant learning relies on the output of the augmented transfer learning, the experiment for assistant learning here indicates that both augmented transfer learning and assistant learning are applied. 

Similar with previous subsection, we conducted experiments for all transfer learning scenarios as in Fig. \ref{fig:transfer_learning_ratios} and Fig. \ref{fig:nt-aug}. The corresponding results are shown in Fig. \ref{fig:al_exp}. In Fig. \ref{fig:al_exp}(c) and (d), the performance of the assistant learning is still similar with its counterpart in Fig. \ref{fig:transfer_learning_ratios} and Fig. \ref{fig:nt-aug}, due to their near-optimal performance. However, in Fig. \ref{fig:al_exp}(a) and (b), we found that the converging time can be further reduced. For example, in Fig. \ref{fig:nt-aug}(a), the augmented transfer learning converges in $125$ trials, while the assistant learning counterpart in Fig. \ref{fig:al_exp}(a) converges in $60$ trials. The experiments for other models can be seen in section \ref{exp_diff_models}.

In summary, by combining the augmented transfer learning and assistant learning, we observe that we can achieve around $1.5 \sim 2.5 \times$ speedup for ResNet20 with superior or comparable pruned accuracy.



\subsection{Experiments for Other DNNs}
\label{exp_diff_models}
In this section, we provide the experiments for other neural networks such as ResNet56, ResNet18, and MobileNet v1. We mainly focused on auto-pruning with a preservation ratio of 0.4 since it is the most challenging case among the 4 cases as we have observed from previous experiments. We put the PocketFlow based baseline, vanilla transfer learning, augmented transfer learning, and assistant learning based auto-pruning in the same table, so that the gain of our proposed framework can be illustrated more clearly. 

Table. \ref{tab:res56} shows the auto-pruning experiments for ResNet56. For transfer learning from ResNet20 with a preservation ratio of 0.4, we observe that vanilla transfer learning, augmented transfer learning and assistant learning have a similar converging speed, which is much faster than the baseline version. $2 \times$ speedup can be achieved. Besides, our proposed assistant learning can achieve higher accuracy. This is because the history based sampling can possibly lead to better design points in the design space. 
For transfer learning from ResNet20 with a preservation ratio of 0.7, we show that our proposed augmented transfer learning and assistant learning have superior converging speed than the vanilla transfer learning counterpart. Our proposed framework can achieve around $1.81 \times$ acceleration over the PocketFlow baseline and $1.36 \times$ acceleration over the vanilla transfer learning counterpart.


Similarly, table \ref{tab:res18} shows the experiments for ResNet18 which has a larger dataset, \emph{e.g.} Imagenet. For transfer learning from ResNet20 with a preservation ratio of 0.4, we observe that both the augmented transfer learning and assistant learning have little performance gain in converging speed and accuracy since the vanilla transfer learning is already close to optimal performance. However, for transfer learning from ResNet20 with a preservation ratio of 0.7, our proposed framework can achieve higher performance in accuracy. This is because transfer learning from "res20 0.7" is more challenging than "res20 0.4", therefore, our proposed learning framework can unleash its superior performance compared with vanilla transfer learning. 

\begin{table}[t]
    \centering
    \caption{Boosting the convergence of ResNet18 with preservation ratio of 0.4}
    \begin{tabular}{|c|c|c|c|}
         \hline
         pruning scenarios & source model & convergence time (trials) & accuracy \\
         \hline
         baseline \cite{wu2018pocketflow} & None & 110 & 0.5063 \\
         \hline
         vanilla transfer & res20 0.4 & 50 & 0.5371 \\
         \hline 
         augmented transfer & res20 0.4 & 50 & 0.5486 \\ 
         \hline
         assistant Learning & res20 0.4 & \textbf{50} & 0.5671 \\
         \hline
         vanilla transfer & res20 0.7 & 60 & 0.4979 \\ 
         \hline 
         augmented transfer & res20 0.7 & 60 &  0.4925 \\ 
         \hline
         assistant learning & res20 0.7 & 60 &  \textbf{0.5704} \\ 
         \hline
    \end{tabular}
    \label{tab:res18}
\end{table}

\begin{table}[t]
    \centering
    \caption{Boosting the convergence of MobileNet v1 with preservation ratio of 0.4}
    \begin{tabular}{|c|c|c|c|}
         \hline
         pruning scenarios & source model & convergence time (trials) & accuracy \\
         \hline
         baseline \cite{wu2018pocketflow} & None & 95 &  0.6443\\
         \hline
         vanilla transfer & res20 0.4 & 60 & 0.6022 \\
         \hline 
         augmented transfer & res20 0.4 & 40 &  0.6136\\ 
         \hline
         assistant Learning & res20 0.4 & \textbf{40} &  0.6673\\
         \hline
         vanilla transfer & res20 0.7 & 50 &  0.6293\\ 
         \hline 
         augmented transfer & res20 0.7 & 50 &  0.6497 \\ 
         \hline
         assistant learning & res20 0.7 & 50 &  \textbf{0.7305} \\ 
         \hline
    \end{tabular}
    \label{tab:mobilenet}
\end{table}


In table \ref{tab:mobilenet}, we show the experiments for MobileNet v1, which is also trained on ImageNet. A time acceleration of $ 2.375 \times$ can be achieved compared with the PocketFlow baseline. A significant accuracy gain in assistant learning can also be observed. 


In summary, in this section, we conducted comprehensive experiments for different preservation ratios and DNN models. A speedup of $1.81 \sim 2.375 \times$ can be observed for different auto-pruning scenarios. A substantial performance gain in accuracy is also observed.

\section{Conclusion}
\label{conclusion}
In this paper, we propose a comprehensive transfer learning framework for the RL agent. An augmented transfer learning and an assistant learning algorithm are proposed to take advantage of the historical data from other pruning scenarios to boost the convergence speed of the network inside the pruning agent, thus saving computing resources and time. The experiments have shown that our framework can significantly reduce the convergence time with superior or comparable pruning accuracy. In the future, we would like to extend our framework to applications like network architecture search, auto quantization, and so on.

\bibliographystyle{IEEEtran}
\begingroup
    \linespread{1}\selectfont
    \bibliography{reference_full}
\endgroup
\end{document}